# Predicting Agricultural Commodities Prices with Machine Learning: A Review of Current Research

Nhat-Quang Tran[†]
School of Science,
Engineering, and Technology,
RMIT University Vietnam
quang.tran26@rmit.edu.vn

Thanh Nguyen Ngoc
School of Science,
Engineering, and Technology,
RMIT University Vietnam
thanh.nguyenngoc@rmit.edu.vn

Quang Tran
School of Science,
Engineering, and Technology,
RMIT University Vietnam
quang.tran@rmit.edu.vn

Anna Felipe
School of Science,
Engineering, and Technology,
RMIT University Vietnam
anna.felipe@rmit.edu.vn

Tom Huynh
School of Science,
Engineering, and Technology,
RMIT University Vietnam
tom.huynh@rmit.edu.vn

Arthur Tang
School of Science,
Engineering, and Technology,
RMIT University Vietnam
arthur.tang@rmit.edu.vn

Thuy Nguyen
School of Science, Engineering, and Technology, RMIT University Vietnam
thuy.nguyen43@rmit.edu.vn

## ABSTRACT

Agricultural price prediction is crucial for farmers, policymakers, and other stakeholders in the agricultural sector. However, it is a challenging task due to the complex and dynamic nature of agricultural markets. Machine learning algorithms have the potential to revolutionize agricultural price prediction by improving accuracy, real-time prediction, customization, and integration.

This paper reviews recent research on machine learning algorithms for agricultural price prediction. We discuss the importance of agriculture in developing countries and the problems associated with crop price falls. We then identify the challenges of predicting agricultural prices and highlight how machine learning algorithms can support better prediction. Next, we present a comprehensive analysis of recent research, discussing the strengths and weaknesses of various machine learning techniques. We conclude that machine learning has the potential to revolutionize agricultural price prediction, but further research is essential to address the limitations and challenges associated with this approach.

[†] Corresponding author

## 1 Introduction

Agriculture is a vital sector that plays a crucial role in the economic development of countries. It provides a source of livelihood for a significant portion of the population [1, 2], especially in developing countries, where it is often the backbone of the economy. Agriculture contributes to employment, income generation, and food security for those countries [3-5]. It also provides the raw materials for a variety of industries, including food processing, textiles, and biofuels [6, 7].

Agricultural commodity price instability can harm a country's GDP and cause emotional and financial distress to farmers who have invested years of effort [8]. Price forecasting can help the agriculture supply chain make informed decisions and mitigate the risks of price fluctuations [9]. By predicting future prices, farmers and other stakeholders can adjust their production and marketing strategies, accordingly, leading to better outcomes for all parties involved. Thus, accurate agricultural product price prediction is essential [8, 10, 11].

Despite its great potential, agriculture price prediction is challenging due to the complex and dynamic nature of the agricultural market, which is influenced by a wide range of factors, including weather variability, supply and demand dynamics, market interdependencies, data availability, and the complexities of agrarian systems [12-20].

Machine learning algorithms have the potential to revolutionize agricultural price prediction [21, 22] by improving accuracy, real-time prediction, customization, and integration. In this study, we systematically review the state-of-the-art research in agriculture price prediction. We systematize the themes, common problems, approaches, and current progress. Based on that, we identify and recommend future research directions.

## 2 Research methodology

We followed the Preferred Reporting Items for Systematic Reviews and Meta-Analyses (PRISMA) framework for data collection and analysis [23]. PRISMA is a guideline that helps systematic reviewers report the purpose, methodology, and findings of their reviews [23-26].

We conducted an exhaustive search of scholarly works in the three most common indexes: Scopus, Web of Science, and Google Scholar. Our search criteria included several combinations of the following keywords: "agriculture", "commodity", "price", "prediction", "machine learning", and "forecasting". This resulted in 110 papers that satisfied the search conditions.

We then reviewed the abstract of each paper to remove irrelevant ones, since with these search terms, we also got papers about, for example, yeild prediction, crop selection, smart farming, and plant disease detection. We also focused on plant rather than animal products, as excluding animal products can significantly contribute to sustainable argiculture by reducing food's land use, greenhouse gases emission and freshwater withdrawals [27]. Additionally, shifting from livestock to plant-based products is humanity for animals and tremendously benefits global food security [28].

We removed articles older than two years that have fewer than nine citations, as this suggests that they are likely not state-of-the-art or high-impact work in the field. However, we kept recent papers (published in 2022 or 2023) regardless their number of citations. Withdrawn or retracted papers are also eliminated. These screening steps lelf us with 27 papers to examine more deeply. The entire list of papers can be found in Table 1.

## 3 Findings

To provide a holistic view of the current state of agricultural price prediction research, we classified all the selected

papers (listed in Table 1) by various criteria, including

1. Country/region
2. Commodity
3. Algorithm

The first criterion, "**Country/region**", sheds light on the context of the research as well as the interest of countries/regions in agricultural price research. An article is counted for a country/region if it either analyzes that country/region's data or if the author is affiliated with that country/region (see Table 2). In the second criterion, "**Commodity**", we show our case data by agricultural products, such as corn, rice, wheat, and soybean. This is important to understand which types of argicultural products received more research (see Table 3). Finally, we classified the articles based on the "**Algorithm**" used. The applied meachine teachniques are varied and include both traditional algorithms and deep learning methods (see Table 4). By analyzing the classification of agricultural price prediction research according to these criteria, we discovered the following findings.

*Finding #1: Current agricultural price prediction research is mainly focused on or by developing countries.*

Table 2: Classification of the reviewed articles by Country/region of analyzed data or authors' affiliation. The 2nd column is the number of articles in that country/region (sorted in descending order). The next column shows the year range in which the articles were published. The last column is the IDs of the published articles (as shown in Table 1).

| COUNTRY | NO. OF ARTICLES | YEARS OF PUBLICATION | ARTICLE IDS |
| --- | --- | --- | --- |
| INDIA | 9 | 2020 - 2023 | [9, 10, 15, 16, 17, 20, 23, 24, 27] |
| CHINA | 6 | 2018 - 2023 | [1, 7, 8, 13, 14, 16] |
| THAILAND | 3 | 2018 - 2022 | [2, 3, 19] |
| UNITED STATES | 3 | 2022 | [18, 21, 22] |
| BANGLADESH | 2 | 2020, 2022 | [5, 12] |
| BRAZIL | 2 | 2019, 2020 | [4, 6] |
| AUSTRIA | 1 | 2022 | [25] |
| DENMARK | 1 | 2022 | [22] |
| FRANCE | 1 | 2022 | [22] |
| GERMANY | 1 | 2022 | [22] |
| ITALY | 1 | 2022 | [25] |
| RUSSIA | 1 | 2021 | [11] |
| SINGAPORE | 1 | 2022 | [25] |

Table 2 shows a list of countries/regions are their number of published papers on agricultural price prediction. It clearly indicates that the majority of research in this subject focuses on developing country contexts, such as India, China, Bangladesh, Brazil and Thailand.

Table 1: List of selected articles sorted by year of publication. The "Country/region" column is the location of the analyzed data or authors' affiliation. Column "Commodity" lists agricultural products whose prices are predicted in the article using the algorithms listed in the "Algorithm" column. "ID" column shows the articles' ID using in Tables 2, 3, and 4. Full names of the algorithms are listed in Table 5.

| ID | AUTHORS | TITLE | YEAR | SOURCE TITLE | COUNTRY/REGION | COMMODITY | ALGORITHM |
|---|---|---|---|---|---|---|---|
| 1 | Xiong T., Li C., Bao Y. | Seasonal forecasting of agricultural commodity price using a hybrid STL and ELM method: Evidence from the vegetable market in China | 2018 | Neurocomputing | China | Cabbage, pepper, cucumber, bean, tomato | STL-ELM |
| 2 | Inyaem U. | Construction Model Using Machine Learning Techniques for the Prediction of Rice Produce for Farmers | 2018 | 3rd IEEE Int. Conf. on Image, Vision and Computing, ICIVC 2018 | Thailand | Rice | DT, ANN |
| 3 | Gertphol S., Chulaka P., Changmai T. | Predictive models for lettuce quality from internet of things-based hydroponic farm | 2018 | 22nd Int. Computer Science and Engineering Conf., ICSEC 2018 | Thailand | Lettuce | SVR, MLR, ANN |
| 4 | Ribeiro M.H.D.M., Ribeiro V.H.A., Reynoso-Meza G., Coelho L.D.S. | Multi-Objective Ensemble Model for Short-Term Price Forecasting in Corn Price Time Series | 2019 | Proceedings of the Int. Joint Conf. on Neural Networks | Brazil | Corn | MOEM |
| 5 | Mehedi Hasan M., Zahara M.T., Mahamudunnobi Sykot M., Hafiz R., Saifuzzaman M. | Solving Onion Market Instability by Forecasting Onion Price Using Machine Learning Approach | 2020 | Int. Conf. on Computational Performance Evaluation, ComPE 2020 | Bangladesh | Onion | KNN, NB, DT, ANN, SVR |
| 6 | Ribeiro M.H.D.M., dos Santos Coelho L. | Ensemble approach based on bagging, boosting and stacking for short-term prediction in agribusiness time series | 2020 | Applied Soft Computing Journal | Brazil | Soybean, wheat | RF, GB, XGB, stacking |
| 7 | Zhang D., Chen S., Liwen L., Xia Q. | Forecasting Agricultural Commodity Prices Using Model Selection Framework with Time Series Features and Forecast Horizons | 2020 | IEEE Access | China | Corn, bean | ANN, SVR, ELM |
| 8 | Yuan C.Z., Ling S.K. | Long Short-Term Memory Model Based Agriculture Commodity Price Prediction Application | 2020 | ACM Int. Conf. Proceeding Series | China | Tomato, chili | ARIMA, LSTM, SVR, Prophet, XGB |
| 9 | Sabu K.M., Kumar T.K.M. | Predictive analytics in Agriculture: Forecasting prices of Arecanuts in Kerala | 2020 | Procedia Computer Science | India | Areca nut | SARIMA, Holt-Winter, LSTM |
| 10 | Paul R.K., Garai S. | Performance comparison of wavelets-based machine learning technique for forecasting agricultural commodity prices | 2021 | Soft Computing | India | Tomato | ANN with wavelet transformation |
| 11 | Tatarintsev M., Korchagin S., Nikitin P., Gorokhova R., Bystrenina I., Serdechnyy D. | Analysis of the forecast price as a factor of sustainable development of agriculture | 2021 | Agronomy | Russia | Sugar | ARIMA |
| 12 | Sarkar J.P., Raihan M., Biswas A., Hossain K.A., Sarder K., Majumder N., Sultana S., Sana K. | Paddy Price Prediction in the South-Western Region of Bangladesh | 2022 | Lecture Notes in Networks and Systems | Bangladesh | Rice | LR, ANN |
| 13 | Liu D., Tang Z., Cai Y. | A Hybrid Model for China's Soybean Spot Price Prediction by Integrating CEEMDAN with Fuzzy Entropy Clustering and CNN-GRU-Attention | 2022 | Sustainability (Switzerland) | China | Soybean | CNN, GRU, Attention |
| 14 | Mao L., Huang Y., Zhang X., Li S., Huang X. | ARIMA model forecasting analysis of the prices of multiple vegetables under the impact of the COVID-19 | 2022 | PLoS ONE | China | Cabbage, carrot, eggplant | ARIMA |
| 15 | Malhotra M., Gupta P., Dhaka V.S., Singh A. | Model to Predict the Crop Cost Using Machine Learning and Read Dataset | 2022 | Lecture Notes in Electrical Engineering | India | Maize, wheat, gram, rice | LR |
| 16 | Babu K.S., Mallikharjuna Rao K. | Onion Price Prediction Using Machine Learning Approaches | 2022 | Lecture Notes on Data Engineering and Communications Technologies | India | Onion | SVR, RF, NB, DT, ANN |
| 17 | Janrao S., Shah D. | Return on investment framework for profitable crop recommendation system by using optimized multilayer perceptron regressor | 2022 | IAES Int. Journal of Artificial Intelligence | India | Cotton, corn | SMO, ANN, Bagging, GR, RF, AdaBoost |
| 18 | Xu X., Zhang Y. | Soybean and Soybean Oil Price Forecasting through the Nonlinear Autoregressive Neural Network (NARNN) and NARNN with Exogenous Inputs (NARNN–X) | 2022 | Intelligent Systems with Applications | United States | Soybean, soybean oil | NARNN, NARNN-X |
| 19 | Tepdang S., Ponprasert R. | Forecasting and Clustering of Cassava Price by Machine Learning (A study of Cassava prices in Thailand) | 2022 | Indonesian Journal of Electrical Engineering and Informatics | Thailand | Cassava | SVR |
| 20 | Thapaswini G., Gunasekaran M. | A Methodology for Crop Price Prediction Using Machine Learning | 2022 | 2022 IEEE 2nd Int. Conf. on Mobile Networks and Wireless Communications, ICMNWC 2022 | India | Rice, wheat, sugarcane, | DT, Neuroevolution |
| 21 | Zhou J., Ye J., Ouyang Y., Tong M., Pan X., Gao J. | On Building Real Time Intelligent Agricultural | 2022 | IEEE 8th Int. Conf. on Big Data Computing Service and Applications, BigDataService 2022 | United States | Corn, oats, soybean, soybean oil | LSTM, GRU, SARIMA |
| 22 | Bonato M., Çepni O., Gupta R., Pierdzioch C. | El Niño, La Niña, and forecastability of the realized variance of agricultural commodity prices: Evidence from a machine learning approach | 2022 | Journal of Forecasting | United States, France, Denmark, Germany | Soybean, soybean oil corn, cocoa, cotton, coffee, oats, rice, sugar, wheat | RF, HAR-RV |
| 23 | Tanwar P., Shah R., Shah J., Lokhande U. | Cotton Price Prediction and Cotton Disease Detection Using Machine Learning | 2022 | Lecture Notes on Data Engineering and Communications Technologies | India | Cotton | LSTM, CNN |
| 24 | Meeradevi, Yasaswi I.G.S., Mundada M.R., Sarika D., Shetty H. | Hybrid Decision Support System Framework for Enhancing Crop Productivity Using Machine Learning | 2022 | Lecture Notes in Networks and Systems | India | Rice, wheat, maize | ARIMA, LSTM |
| 25 | Hammad A.T., Falchetta G. | Probabilistic forecasting of remotely sensed cropland vegetation health and its relevance for food security | 2022 | Science of the Total Environment | Singapore, Austria, Italy | Rice | Quantile RF |
| 26 | Zeng L., Ling L., Zhang D., Jiang W. | Optimal forecast combination based on PSO-CS approach for daily agricultural future prices forecasting | 2023 | Applied Soft Computing | China | Corn, wheat | ARIMA, ETS, ANN, ELM |
| 27 | Murugesan G., Radha B. | An extrapolative model for price prediction of crops using hybrid ensemble learning techniques | 2023 | Int. Journal of Advanced Technology and Engineering Exploration | India | Coconut | ARIMA, SVR |

This can be attributed to the significant role of agriculture as a major sector in the economies of these countries [29, 30]. However, agricultural price prediction is also crucial in government tasks such food's land-use and crop planing [31] which are done well in developed countries. Hence, we urge developed countries to be more active in publishing and sharing research results on this topic which could greatly benefit farmers, policy markers, and other stakeholders in the agricultural sector around the world, especially in developing countries.

*Finding #2: The absence of systematic research of price prediction for agricultural product groups.*

Table 3: Classification of the reviewed articles based on product. The table is sorted by the column "NO. OF ARTICLES" (descending order). Article's IDs are shown in Table 1.

| COMMODITY | NO. OF ARTICLES | YEARS OF PUBLICATION | ARTICLE IDS |
|---|---|---|---|
| CORN/MAIZE | 8 | 2019 - 2023 | [4, 7, 15, 17, 21, 22, 24, 26] |
| RICE | 7 | 2018, 2022 | [2, 12, 15, 20, 22, 24, 25] |
| WHEAT | 6 | 2020, 2023 | [6, 15, 20, 22, 24, 26] |
| SOYBEAN | 5 | 2020, 2022 | [6, 13, 18, 21, 22] |
| TOMATO | 3 | 2018 - 2021 | [1, 8, 10] |
| SOYBEAN OIL | 3 | 2022 | [18, 21, 22]] |
| COTTON | 3 | 2022 | [17, 22, 23] |
| CABBAGE | 2 | 2018, 2022 | [1, 14] |
| BEAN | 2 | 2018, 2020 | [1, 7] |
| OATS | 2 | 2022 | [21, 22] |
| SUGAR | 2 | 2021, 2022 | [11, 22] |
| ARECA NUT | 1 | 2020 | [9] |
| CARROT | 1 | 2022 | [14] |
| CASSAVA | 1 | 2022 | [19] |
| EGGPLANT | 1 | 2022 | [14] |
| CUCUMBER | 1 | 2018 | [1] |
| PEPPER | 1 | 2018 | [1] |
| COCONUT | 1 | 2023 | [27] |
| LETTUCE | 1 | 2018 | [3] |
| GRAM | 1 | 2022 | [15] |
| SUGARCANE | 1 | 2020 | [22] |
| COCOA | 1 | 2022 | [22] |
| COFFEE | 1 | 2022 | [22] |
| CHILI | 1 | 2022 | [8] |

Table 3 presents a list of agricultural commodities and their inclusion in research papers related to price prediction. It is evident that research on this topic encompasses a wide range of products [14, 32]. The commodities that receive the most research attention are corn/maize, rice, wheat, and soybean which are all staple food consumed around the world. Additionally, the world's most productive countries of products such corn and wheat are developed countries [33]. This aligns with the Finding #1.

Furthermore, we raise the question regarding the absence of research on systematic reviews of price prediction for agricultural product groups. Since the process of individually analyzing specific products for price prediction can be laborious, grouping crops with similar weather and cultivation conditions and market shares could foster more focused research and better impact of this field [32, 34].

*Finding #3: LSTM and neural network-related approaches are the most used methods.*

Table 4: Classification of the reviewed articles based on algorithms used. The table is sorted by the column "NO. OF ARTICLES" (descending order). Article's IDs are shown in Table 1. Full names of the algorithms are listed in Table 5.

| ALGORITHM | NO. OF ARTICLES | YEARS OF PUBLICATION | ARTICLE IDS |
|---|---|---|---|
| NEURAL NETWORKS* | 16 | 2018 - 2023 | [2, 3, 5, 7, 8, 9, 10, 12, 13, 16, 17, 18, 20, 21, 23, 24, 26] |
| ARIMA/SARIMA | 8 | 2020 - 2023 | [8, 9, 11, 14, 21, 24, 26, 27] |
| SVR | 7 | 2018 - 2023 | [3, 5, 7, 8, 16, 19, 27] |
| ENSEMBLE LEARNING** | 6 | 2020, 2022 | [6, 8,16, 17, 22, 25] |
| DT | 4 | 2018 - 2022 | [2, 5, 16, 20] |
| ELM | 3 | 2018 - 2023 | [1, 7, 26] |
| LR/MLR | 3 | 2018, 2022 | [3, 12, 15] |
| NB | 2 | 2020, 2022 | [5, 16] |
| ETS | 1 | 2023 | [16] |
| PROPHET | 1 | 2020 | [8] |
| KNN | 1 | 2020 | [5] |
| MOEM | 1 | 2019 | [4] |
| HAR-RV | 1 | 2022 | [22] |
| HOLT-WINTER | 1 | 2020 | [9] |
| SMO/ IMPROVED-SMO | 1 | 2022 | [17] |

\* NEURAL NETWORKS includes ANN, CNN, LSTM, GRU, Attention, NARNN, NARNN-X and Neuroevolution.

\*\* ENSEMBLE LEARNING includes Bagging, RF, Quantile RB, GB, XGB, AdaBoost, Stacking.

Table 4 demonstrates the diversity of machine learning algorithms that are used for agriculture price prediction. LSTM and neural network-related approaches prominently feature in this research subject, with a total of 16 articles. Time series algprithms, namely ARIMA and its improved variant SARIMA, are also popular (8 articles) because most of the historical price data is in this format. Tranditional methods such as SVR, DT and emsemble learning are also widely used. Ensemble and hybrid techniques are particularly promising for future development because they combine individual predictors, including newly developed ones, to improve their predictions.

Table 5: List of algorithm names (sorted alphabetically).

| ABBREVIATION | MODEL FULL NAME |
|---|---|
| ANN | Artificial Neural Network |
| ARIMA | Autoregressive Integrated Moving Average |
| ATTENTION | Attention Mechanism |
| CNN | Convolutional Neural Network |
| DT | Decision Tree |
| ELM | Extreme Learning Machine |
| ETS | Exponential Smoothing |
| GB | Gradient Boosting |
| GR | Gaussian Regressor |
| GRU | Gated Recurrent Unit |
| HAR-RV | Heterogeneous Autoregressive Realized Variance |
| KNN | K-Nearest Neighbor |
| LR | Linear Regression |
| LSTM | Long Short-Term Memory |
| MLR | Multiple Linear Regression |
| MOEM | Multi-Objective Ensemble Model |
| NARNN | Nonlinear Autoregressive Neural Network |
| NARNN–X | NARNN With Exogenous Inputs |
| NB | Naïve Bayes |
| RF | Random Forest |
| SMO | Sequential Minimal Optimization Regressor |
| STL | Seasonal-Trend Decomposition Procedures Based on Loess |
| SVR | Support Vector Regression |
| XGB | eXtreme Gradient Boosting |

## 4 Discussion

### 4.1 Where are you now?

Our findings show that significant progress is being made in agricultural price prediction research. Researchers are using various machine learning algorithms to improve the accuracy and reliability of agricultural price forecasts. One notable area of advancement is the utilization of hybird or emsemble of machine learning techniques uncluding both traditional methods such as SRV, DT and deep learning approaches such as LSTM, GRU, Attention. These algorithms have shown promising results in capturing complex patterns and nonlinear relationships in agriculture price data.

Researchers are also incorporating additional factors and variables into their models to enhance prediction accuracy [35-38]. These factors may include weather conditions, market trends, supply and demand dynamics, and socio-economic factors [6, 12, 37, 39]. By integrating these variables, researchers aim to create more comprehensive and robust models for agriculture price prediction. Another area of progress is the integration of big data and advanced analytics techniques. With the availability of vast amounts of data, including historical price data, market reports, and satellite imagery, researchers can leverage data-driven approaches to gain deeper insights into price patterns and make more accurate predictions [40].

### 4.2 What are future directions?

Our findings suggest that collaborative efforts and interdisciplinary research are needed in the field of agriculture price prediction, as current research is mainly focused on or by developing countries. Although Western and developed countries likely have extensive advanced knowledge and techniques in this field [14], they are not well-represented in the current state of price prediction research. Another area for future exploration would be a more streamlined approach in terms of algorithms, methodologies as well as crop groups.

Finally, we call for more exploration of the integration of advanced technologies, such as remote sensing, Internet of Things (IoT), and blockchain, into agriculture price prediction [41]. Utilize satellite imagery, weather data, fuel price data, and IoT sensors to gather real-time information and incorporate it into prediction models. Explore the potential of blockchain technology to enhance transparency, traceability, and trust in agricultural supply chains, which could ultimately impact price prediction accuracy [14, 42].